\documentclass[journal]{IEEEtran}

\usepackage[utf8]{inputenc}

\ifCLASSINFOpdf
\else
\fi
\usepackage{hyperref}
\usepackage{xcolor}
\usepackage{graphicx}
\usepackage{subfigure}
\usepackage{amsfonts}
\usepackage{amsmath}
\usepackage{booktabs}
\usepackage[utf8]{inputenc}
\usepackage{tabularx}

\begin{document}

\title{Deep Sequential Models for Suicidal Ideation from Multiple Source Data 
\thanks{
Manuscript received December 17, 2018. This work has been funded in part by the Spanish MINECO under Grants TEC2015-69868-C2-1-R, TEC2016-78434-C3-3-R and TEC2017‐92552‐EXP, in part by Spanish MICINN under Grant RTI2018-099655-B-I00, in part by Comunidad de Madrid under Grants IND2017/TIC-7618, IND2018/TIC-9649, Y2018/TCS-4705 and B2017/BMD-3740 AGES-CM 2CM, in part by BBVA Foundation under 2018 Grant for Deep-DARWiN project, in part by ISCIII under Grant PI16/01852 and in part by AFSP under Grant LSRG-1-005-16. We gratefully acknowledge the support of NVIDIA Corporation with the donation of the Titan X Pascal GPU used for this research. (Corresponding author: Ignacio Peis).}
}

\author{Ignacio Peis, Pablo M. Olmos, Constanza Vera-Varela, María Luisa Barrigón, Philippe Courtet, Enrique Baca-García and Antonio Artés-Rodríguez, ~\IEEEmembership{Senior Member,~IEEE.} 
\thanks{I. Peis, P. M. Olmos and A. Artés-Rodríguez are with the Dept. of Signal Theory and Communications, UC3M, Leganés, Madrid, 28911 Spain and also with the Health Research Institute Gregorio Marañón, 28007 Madrid e-mails: (ipeis@tsc.uc3m.es; olmos@tsc.uc3m.es; antonio@tsc.uc3m.es).}
\thanks{C. Vera-Varela, M. L. Barrigón and E. Baca are with the Dept. of Psychiatry, IIS-Jimenez Díaz Foundation, Madrid, 28040 Spain e-mails: cveravarela@gmail.com; luisa.barrigon@fjd.es; ebacgar2@yahoo.es.}
\thanks{P. Courtet is with the Dept. of Psychiatric Emergency and Acute Care, Lapeyronie Hospital, University of Montpellier, Montpellier, 34295 France e-mail: p‐courtet@chu‐montpellier.fr.}
\thanks{M. L. Barrigón and E. Baca are also with the Dept. of Psychiatry, Universidad Autónoma de Madrid, Madrid, 28049 Spain.}
\thanks{A. Artés-Rodríguez and E. Baca-García are also with the CIBERSAM, ISCIII, Madrid, 28029 Spain.}
\thanks{E. Baca-García is also with Universidad Católica del Maule, Talca, Chile.}

}

\markboth{ P\MakeLowercase{ublished in} IEEE Journal of Biomedical and Health Informatics. DOI 10.1109/JBHI.2019.2919270 }%
{IEEE Journal of Biomedical and Health Informatics}

\maketitle

\begin{abstract}
This article presents a novel method for predicting suicidal ideation from Electronic Health Records (EHR) and Ecological Momentary Assessment (EMA) data using deep sequential models. Both EHR longitudinal data and EMA question forms are defined by asynchronous, variable length, randomly-sampled data sequences. In our method, we model each of them with a Recurrent Neural Network (RNN), and both sequences are  aligned by concatenating the hidden state of each of them using temporal marks. Furthermore, we incorporate attention schemes to improve performance in long sequences and time-independent pre-trained schemes to cope with very short sequences. Using a database of 1023 patients, our experimental results show that the addition of EMA records boosts the system recall to predict the suicidal ideation diagnosis from 48.13\% obtained exclusively from  EHR-based state-of-the-art methods to 67.78\%. Additionally, our method provides interpretability through the t-SNE representation of the latent space. Further, the most relevant input features are identified and interpreted medically.
\end{abstract}


\IEEEpeerreviewmaketitle


\section{Introduction}

\IEEEPARstart{A}{lthough} research on suicidal behavior has intensified in the last decade and there is now a greater understanding of the factors at play, suicide remains a complex public health problem \cite{turecki2016suicide} and the global prevalence of suicide continues to be high,  with an annual global age-standardized suicide rate of 11.4 per 100 000 people \cite{intro_who}. Globally, lifetime prevalence rates are approximately 9.2\% for suicidal ideation and 60\% of transitions from ideation to plan and attempt occur within the first year after ideation onset \cite{nock2008cross}.

The vast majority of the studies in suicide are based on self report measures of indicator variables as hopelessness, depressive symptoms, sleep disorders or violence \cite{intro_ema}. These traditional methods for assessment have the disadvantage of considering long window assessments (weeks, months, or years), which might exclude crucial reports about who is at imminent risk for suicidal behavior \cite{czyz2018ecological}. In suicidology, as well as in other areas of psychiatry, technological advances have enhanced ambulatory monitoring techniques and facilitate the use of Ecological Momentary Assessment (EMA) in combination with traditional Electronic Health Records (EHR).  EMA allows to measure variables of interest in real-time and in natural contexts of daily life \cite{barrigon2017user}, eliminating inaccurate reporting of the time elapsed between the prompt and the completion of the questionnaire, and has demonstrated to be safe in suicidal thoughts assessment \cite{husky2014feasibility}. The employment of EMA using smartphones is opening up new perspectives in suicidal ideation studies, making possible the identification of different digital phenotypes of suicidal thoughts \cite{kleiman2018digital,bernanke2017toward}.

In the context of computational psychiatry, there exist a small group of studies that apply machine learning techniques to predict suicidal behavior.  In \cite{intro_pred}, the authors develop hand-crafted patient features from longitudinal EHR data of mental health and primary care visits from a group of nearly 3 million patients to predict suicide attempts and suicide deaths within 90 days, obtaining 0.851 and 0.861 of ROC (Receiver Operating Characteristic), respectively.  In \cite{intro_pred2}, a similar setup is used to predict suicides and suicidal behavior from longitudinal EHR within three to four years in advance (0.45 recall / 0.9 precision). In both cases, suicide prediction is performed using a long-temporal window (3 months and up to 4 years respectively). Combined with a population of patients in which each of them is followed for years (in \cite{intro_pred}, 56\% of the patients in the study are treated for more than 5 years, and in \cite{intro_pred2} the average period is 5.27 years), these results in a well-condition prediction problem, easily addressed by LASSO (Least Absolute Shrinkage and Selection Operator) logistic regression \cite{intro_pred} and naive Bayesian classifiers \cite{intro_pred2}. Clearly, this setup is far from the reality that a clinician faces, as the longitudinal EHR data for each patient is in most cases less than one year and a half. In this study, the average periods of the treatments are 1.3 $\pm$ 0.86 years for EHR and 0.33 $\pm$ 0.44 years for EMA.

In this work, we develop sequential models based on neural-networks \cite{Lipton2015} for predicting the identification of suicidal ideation, identified by the clinician during the presential evaluation at every follow-up visit of the patient. The difficulty arises in modeling two sources of asynchronous, length-variable and randomly-sampled sequences, i.e. EHR and EMA. 
The random nature of these sequences w.r.t. sampling and length arises in the fact that they are obtained from patients appointments with the clinician, which are in general not periodic. Also, the patient's pattern in accessing EMA mobile application is in general completely random. 

We develop Recurrent Neural Networks (RNN) models able to combine patient's EHR longitudinal data along with the EMA question forms. By combining pre-trained feed-forward Neural Networks (NNs) and attention models \cite{mod_frustr}, our model learns from the heterogeneous sequences in a fair basis, preventing that the model training overfits on long temporal sequences for which more information is available.  Using a test patient population, we show our method is able to reach a 0.67 recall performance, while, when using only EHR longitudinal data, it downs to 0.46, which is comparable to EHR-based state-of-the-art methods \cite{intro_pred}-\cite{intro_pred2} that do not explicitly model temporal dependencies. This demonstrates the great impact that EMA can have in patient assessment. Further, our model only effectively captures the EMA influence on the diagnosis when is combined with neural attention models \cite{bahdanau2014neural}. Therefore, from both the technical method proposed and the application sides, we believe this work represents a step forward in the use of machine-learning methods in suicidology. 

The paper is organized as follows. Firstly, Section \ref{sec:problem_statement} introduces the problem we are facing with (\ref{sec:case_definition}) and the solution we propose (\ref{sec:method_proposed}). In Section \ref{design} we expose an in-depth presentation of the model architecture. Section \ref{sec:experiments} shows numerical results and finally Section \ref{sec:discussion} contains some concluding remarks.


\section{Problem Statement}\label{sec:problem_statement}

In this work we aim at predicting the identification of suicidal ideation by a mental health clinician (psychiatrist or clinical psychology) using multiple information sources that contain relevant information about the evolution of a patient along his/her treatment. On one hand, we use EHR filled up by the clinicians at every follow-up visit with the patient, which we call \emph{clinician sequence}. For the EMA data, the so-called \emph{patient sequence}, we use \emph{MEmind Wellness Tracker}\footnote{The web application is available at \url{www.memind.net} using internet‐connected devices with any operating system.}, a web application for mobile devices developed to merge different data sources and provide summaries of the patient state in clinical practice \cite{barrigon2017user}.

\subsection{Case Definition}\label{sec:case_definition}

The EHR data variables are organized in four categories: 1) sociodemographic information; 2) diagnoses according International Statistical Classification of Diseases and Related Health Problems 10th revision (ICD-10 \cite{organisation1992icd}), questions regarding history of aggression and violence, suicidal thoughts and behavior \cite{hergueta1998mini}, the Clinical Global Impression (CGI) scale that rates the psychiatrist’s impression of the severity of psychopathology and the variation since last visit on a similar seven-point scale \cite{guy1976early}, and the Global Assessment of Functioning (GAF) Scale \cite{spitzer1996global};  3) Brugha’s life threatening events scale \cite{brugha1985list} and 4) collecting treatment.

The EMA data variables belong to three categories 1) the \textit{How are you today?}, a clinical-based questionnaire; 2) the WHO-5 Well-Being Index (World Health Organization \cite{who5}),  and 3) the 12-Item General Health Questionnaire (GHQ-12) \cite{sanchez200812}. 

Note that the EHR view also contains the variable we aim at predicting, i.e.,  whether the clinician considers that patient presents \emph{suicidal ideation} or not. This variable is not included in the \emph{clinician sequence}, as it is the target to be predicted at every visit. 

\begin{figure}[h]
	\centering
	\includegraphics[width=0.8\linewidth]{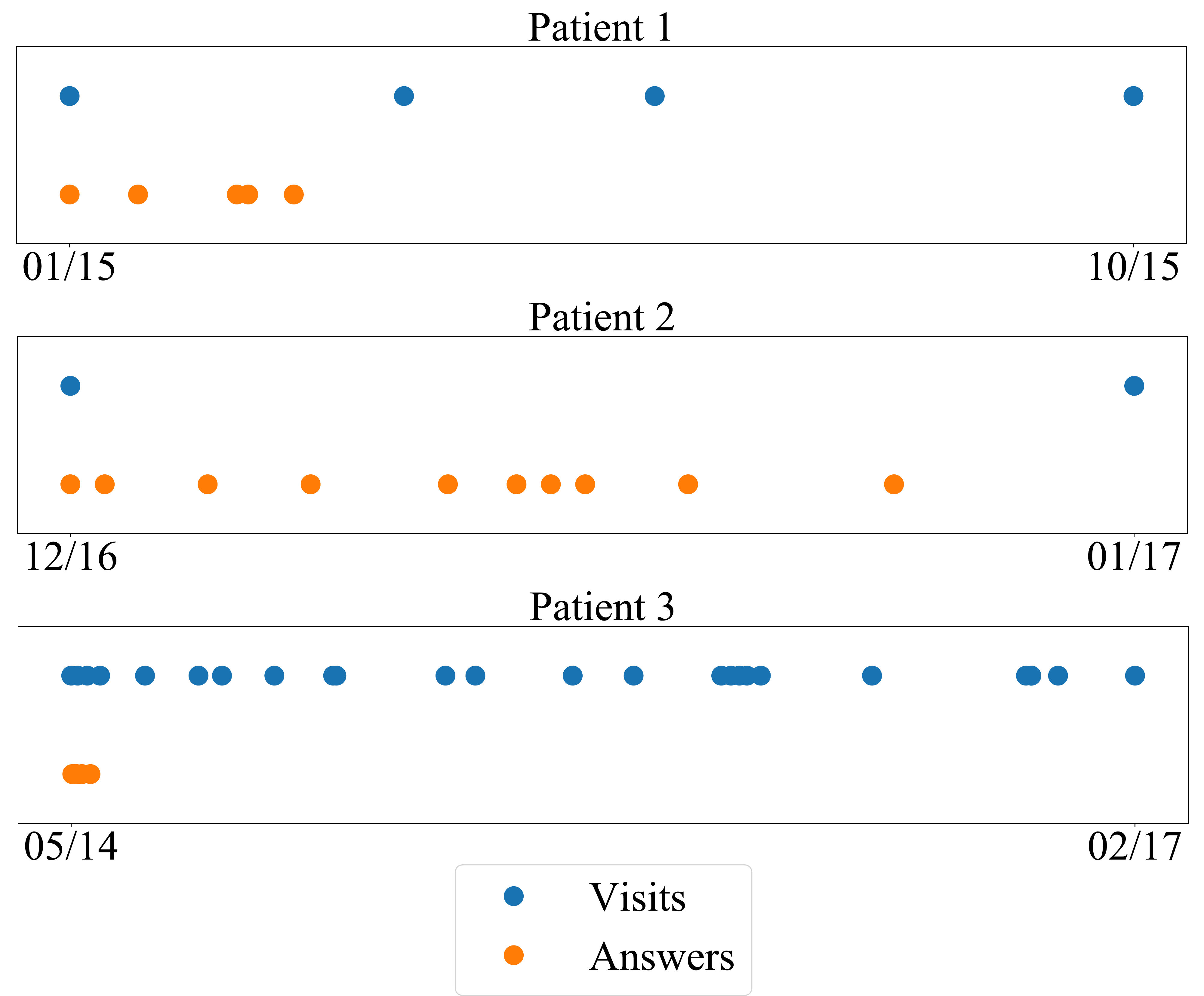}
	\caption{Clinician and patient sequences associated to three different patients.}
	\label{fig:sequences}
\end{figure}

We have collected a database formed by the two different sources (EHR and EMA). The patients were evaluated in psychiatric routine or psychological visits at mental health facilities affiliated with the Fundación Jiménez Díaz Hospital in Madrid, Spain. We considered only those patients that accessed the EMA interface on at least three occasions. From May 2014 to Sep 2016, a sample of 1023 patients accomplished this criteria. Participants were mostly women (64.7\%), with a mean age of 43.32 $\pm$ 12.6 years old, an active job status (51.4\%) and concerning civil status, 57.2\% were married. Most patients were diagnosed with anxiety related disorders (F40-F49 ICD-10) (52.6\%) and with mood disorders (F30-F39 ICD-10) (23.3\%).

Temporal patterns are extremely diverse. For instance, in Fig. \ref{fig:sequences} we include an example of the two sequences associated to three patients. Each point corresponds to an observation. The prediction model must be able to learn from variable-length sequences in which the typical sequence length is very short (less than 10 in both cases), but at the same time a certain subset of patients are characterized with very large sequences. The maximum length of the patient sequences is 858, while the minimum is 3. The maximum length of the clinician sequences is 119, while the minimum is only one. The mean lengths are 19.66 answers and  7.87 visits, respectively.

\subsection{Method proposed}\label{sec:method_proposed}

We build a temporal classifier that aims at predicting suicidal ideation at every visit between the clinician and the patient. Unlike most of the studies in the related literature, we do take into account the temporal correlation of the different sequences by using RNNs \cite{Lipton2015, graves_thesis}. A RNN is able to capture signal correlation by sharing the NN weights along time. Deep Learning architectures are able to handle the heterogeneity present in the data and implicitly adapt the variable length of the sequences, as well as capture both short and long-term temporal correlations. In order to effectively model the two asynchronous sequences, we propose to use a different RNN for each of them, where the time index is determined by every follow-up visit in the clinician sequence, and every time the patient fills up a MEmind questionare in the patient sequence.

\begin{figure}[h]
	\centering
	\includegraphics[width=0.98\linewidth]{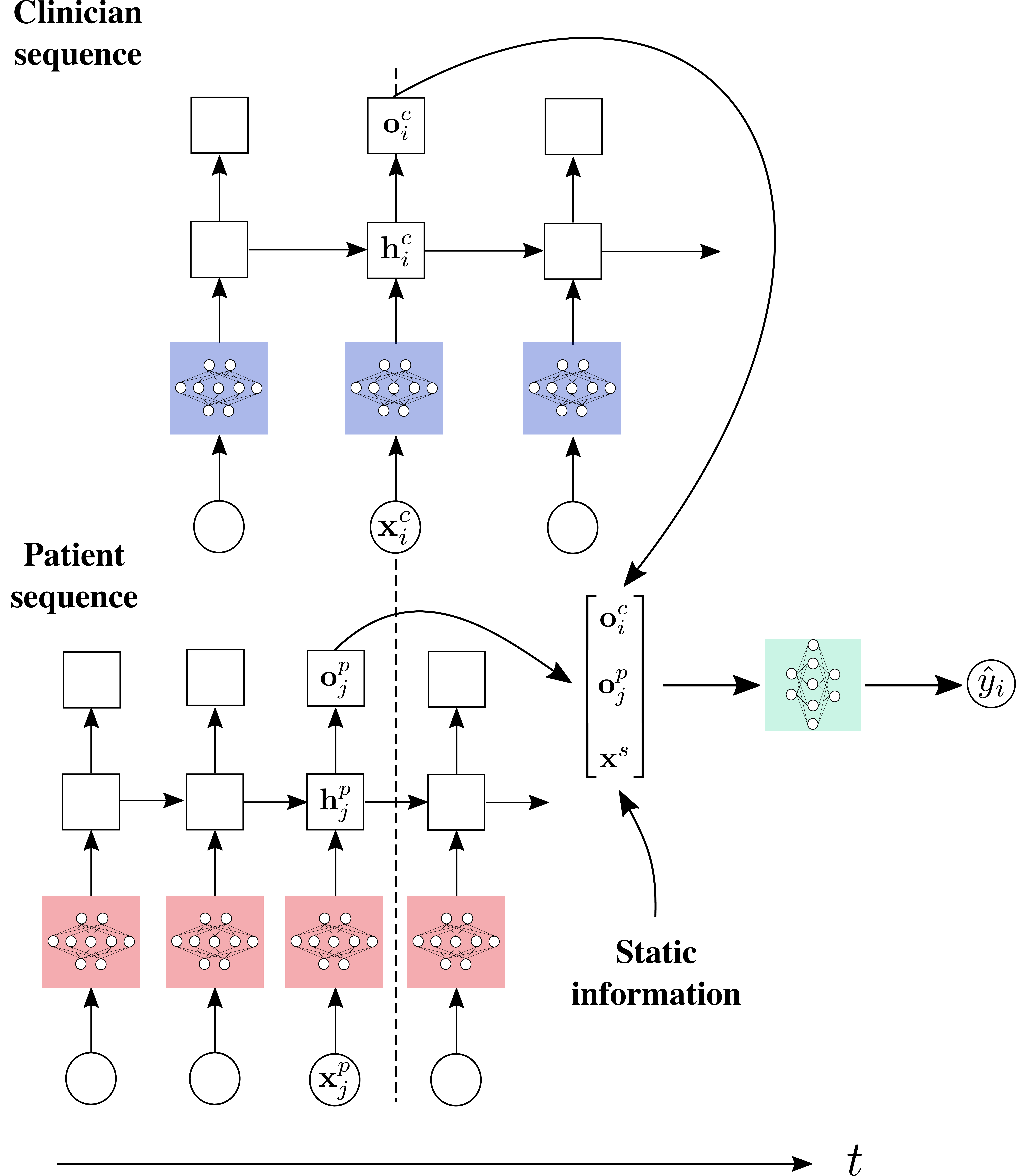}
	\caption{Model architecture, showing the non-linear transformation of the inputs and the the merging of the two sequences, adding additional static information. The figure represents an example of how the model makes a prediction at the $i$-th patient-clinician visit. Circles refer to observations, whereas rectangles refer to states.}
	\label{fig:diag_normal}
\end{figure}

Every event comes with a time stamp that will be used to align both sequences. As illustrated in Fig. \ref{fig:diag_normal}, to predict the target variable at every visit, we concatenate the output of the RNN modeling the clinician sequence with the most-recently output of the RNN modeling the patient sequence, and the resulting vector will be used as input to a final classifier built over a feedforward NN with sigmoidal output activation.  In order to limit the number of parameters and control overfitting, NN layers for dimensionality reduction are employed. A detail description of the model and its optimization will be provided in Section \ref{design}.  

As both the clinician RNN and the patient RNN are driven by events and not by a natural temporal scale, the elapsed time between events will be lost unless we explicitly encode it as input information to the RNN. Despite RNNs naturally handle variable-length sequences, the prediction model is trained over a non-uniform database where more than one-third of the sequences are not really temporal sequences, as their length is only one. As shown later, a naive RNN implementation would not result in accurate results. Furthermore, the use of attention schemes (see Section \ref{sec:att}) is the key to effectively incorporate the patient sequence. In words, attention schemes learn to represent variations in the patient state, which is what certainly clinicians observe, rather than punctual patient MEmind answers.

\section{Design methodology}\label{design}

In this section we present the proposed model, exposed in Figure \ref{fig:diag_normal}. Firstly, we describe our notation. Secondly, every part of the architecture is analyzed.

\subsection{Notation}

We deal with EHR (clinician) and EMA (patient) data sources from a group of $N$ patients. If we use $k_c$ and $k_p$ variables from the clinician and patient data, respectively, the $n$-th patient can be represented by two sequences of $T_n^c$ and $T_n^p$ tuples: $( t_{n, i}^{c}, \textbf{x}_{n, i}^{c}) \in \mathbb{R} \times \mathbb{R}^{k_c} , i=1, ..., T_n^c$ and $( t_{n,j}^{p}, \textbf{x}_{n,j}^{p}) \in \mathbb{R} \times \mathbb{R}^{k_p} , j=1, ..., T_n^p$, where $t_{n,i}^{c}$ and $t_{n,j}^{p}$ denote the time of the $i$-th visit and $j$-th answer of the $n$-th patient. Each $\textbf{x}_{n,i}^{c}$ and $\textbf{x}_{n,j}^{p}$ are multivariate observations. We also include the static (non sequential) information of the patient as $\textbf{x}_n^s$. The goal is to predict the binary label $y_{n,i}$ at each clinician time step $t_{n,i}^{c}$. In the following, we do not include the index of the patient $n$ for simplicity.

\subsection{Sequence Modeling and Dimensionality Reduction}\label{sec:seq_mod}

Provided that we have a fairly small database, and with the aim at avoiding overfitting, we use basic RNNs instead of more complex architectures like Long Short Term Memory (LSTM \cite{hochreiter1997}) or Gated Recurrent Unit (GRUs \cite{cho2014learning}). The RNN is defined by the following recursion \cite{Lipton2015}:
\begin{equation}
    \textbf{h}_i = \sigma \left( W^{hx} \textbf{x}_{i} + W^{hh}\textbf{h}_{i-1}+\textbf{b}_h \right)
\end{equation}
where $\sigma$ is a non linear activation (typically hyperbolic tangent or Rectified Linear Unit), $\textbf{x}_{i}$  the input (in our case, a multivariate observation with index $i$), $\textbf{h}_i$ the so-called RNN state (with index $i$), $W^{hx}$, $W^{hh}$ projection matrices and $\textbf{b}_h$ the bias term. The RNN output is computed as follows:
\begin{equation}
    \hat{\textbf{o}}_i = \text{softmax}(W^{yh}\textbf{h}_i+\textbf{b}_y)
\end{equation}

Further, a dimensionality reduction stage is performed over the input data to both the patient and clinician sequence. To this end, we compared two different solutions. On one hand, we can use a simple linear layer consisting on an embedding matrix \cite{choi2016retain} with dimensions $N\times I$  ($N$ refers to the dimension of the data, $I$ to the number of inputs for the RNN). This matrix is trained along with the rest of parameters of the
structure in Fig. \ref{fig:diag_normal}. On the other hand, we propose to use a nonlinear dimensionality reduction driven by a dense layer with hyperbolic tangent output activation. This helps us to better accommodate very short sequences, as this layer is pre-trained separately over a prediction model in which we neglect temporal correlations (no RNNs are used) and suicidal ideation $y_i$ is predicted directly from the output of these layers, separately for the clinician (using $\textbf{x}_i^c$) and patient (using $\textbf{x}_j^p$) sources. Due to the use of these pre-trained dense layers in Fig. \ref{fig:diag_normal}, the inputs of the RNN are already discriminative, and the RNN has the specific role of capturing temporal correlations.

\subsection{Sequence merging and Cost Function}\label{sec:merge}

Both the clinician and patient RNNs are aligned in time upon prediction. To this end, we concatenate the output vector $\textbf{o}_i^{c}$ of the clinician RNN, the selected output $\textbf{o}_{j}^{p}$ (index $j$ of the most recent answer) of the patient RNN and the static information $\textbf{x}^s$, to form a new vector that is going to feed another feed forward NN that performs the detection (see Fig. \ref{fig:diag_normal}). The static information is a 2-dimesional vector containing a binary feature with the sex of the patient and a scalar feature with his/her age. The classifier consists on a feed-forward NN that applies a sigmoid activation at the last layer to obtain predictions between 0 and 1. As described in Section \ref{sec:experiments}, after cross validating several structures, a simple single dense layer is selected to implement this classifier.

We are using a method similar to the target replication technique exposed in \cite{lipton2015learning}, in which the cost function is given by the sum of the cross entropy function between the probabilities $\hat{y}_{n, i}$ and the target $y_{n, i}$ at each $t_{n,i}^{c}$, following:
\begin{equation}
\mathcal{L}(\boldsymbol{\Theta}) = - \sum_{n=1}^{N} \frac{1}{T_n^c} \sum_{i=1}^{T_n^c} \left( y_{n, i} \log \hat{y}_{n, i} + (1-y_{n, i}) \log (1-\hat{y}_{n, i}) \right)
\end{equation}

\subsection{Learning the initial state of the RNNs}

\begin{figure}[h]
	\centering
	\includegraphics[width=0.5\linewidth]{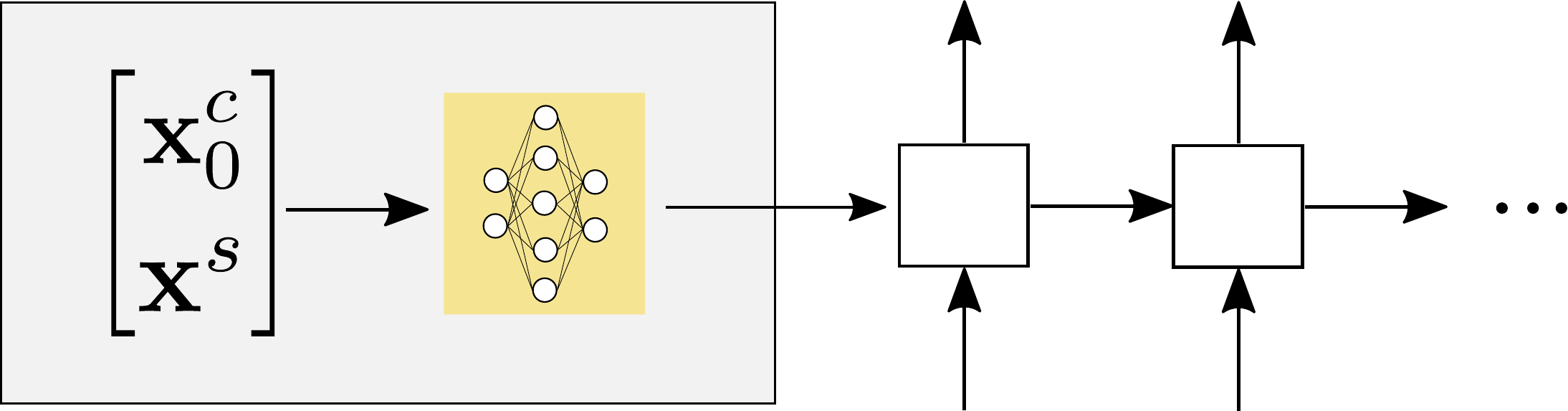}
	\caption{Schematic of the initial state learning method.}
	\label{fig:initial}
\end{figure}

To reduce dependency w.r.t. initialization, we incorporate a feed forward NN that, using the initial vector of the clinician sequence and the patient static information, learns the initial state of the clinician RNN (Figure \ref{fig:initial}), and thus we expect an improved adaptation of the model for short sequences. 

\subsection{Attention schemes}\label{sec:att}

\begin{figure}
	\centering
	\includegraphics[width=0.9\linewidth]{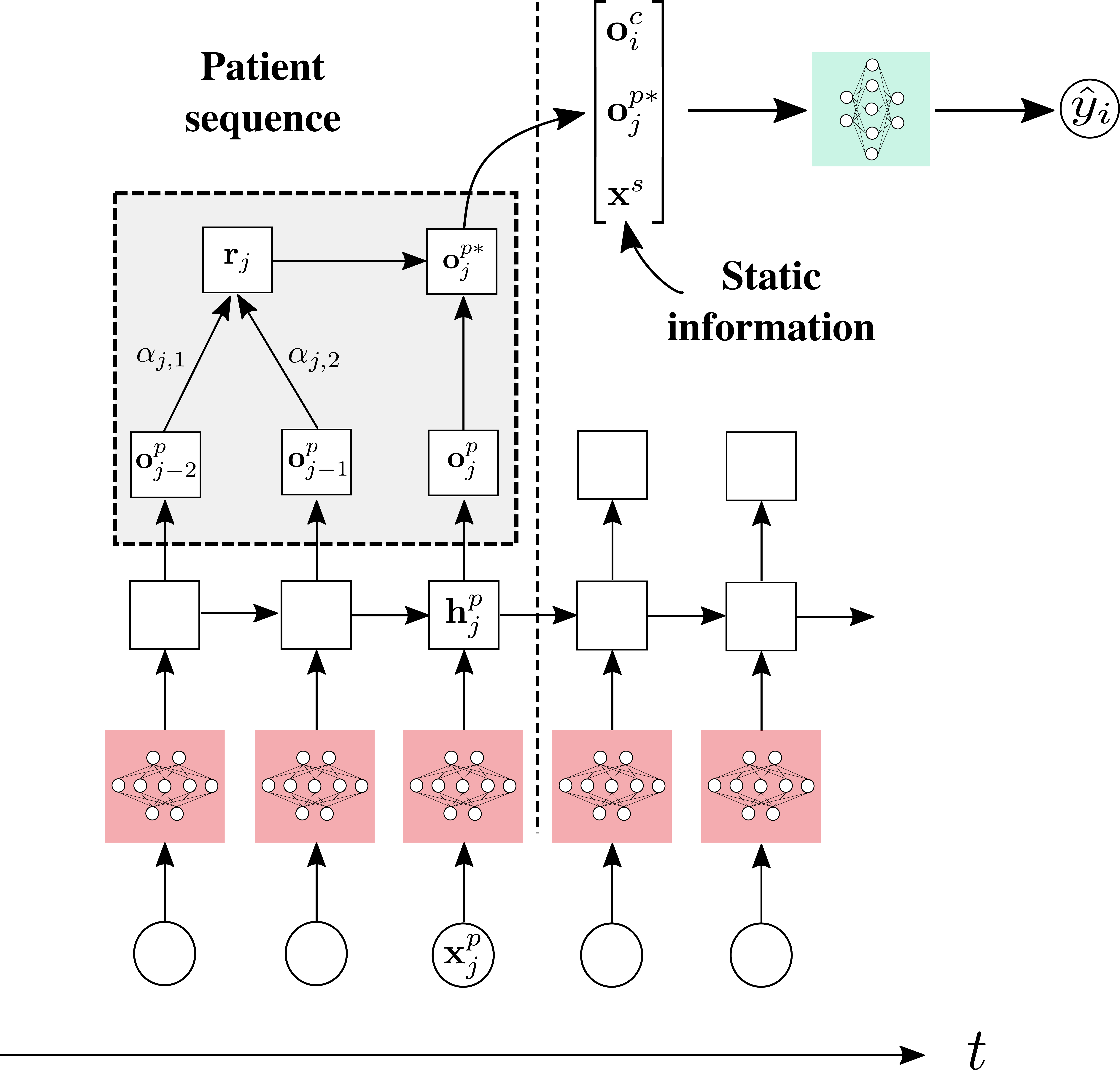}
	\caption{Proposed model architecture, including an attention scheme on the patient sequence. For this graphical example, we use $L=2$.}
	\label{fig:diag_att}
\end{figure}

The use of attention schemes \cite{deng2018latent, kim2017structured} within sequence RNN-based models have become a recent revolution in fields such as Natural Language Processing \cite{mod_frustr}, or computer vision \cite{mnih2014recurrent}. In a nutshell, attention models allow neural sequence models to find discriminative representations in the evolution of the state of the RNN, applying different weights in the more recent history that can facilitate learning mid and long-range dependencies. This idea has been adapted to EHR data in architectures like RETAIN \cite{choi2016retain}, where authors use a two-level neural attention mechanism  with the aim at increasing the model interpretability. We propose a windowed version of attention scheme based on \cite{mod_frustr}, where we can tune the size of the past memory. This architecture will lead us to focus on the most recent answers (see Figure \ref{fig:diag_att}) in the patient sequence, transforming the outputs of the RNN into a new attention-based space. We take the previous $L$ output vectors of the patient's RNN (for sequences shorter than $L$, we apply zero padding) as memory $\textbf{Y}_i = \left[ \textbf{o}_{i-L}  \, ... \, \textbf{o}_{i-1} \in \mathbb{R}^{k\times L} \right] $, where $k$ is the dimension of the outputs. In figure \ref{fig:diag_att} a schematic of the proposed attention schemes is included. We define $\textbf{1} \in \mathbb{R}^L$ as a vector of ones. Then, as in \cite{mod_frustr}, we use:
\begin{align}
& \textbf{M}_i = \tanh \left( \textbf{W}^Y \textbf{Y}_i  + \left( \textbf{W}^o \textbf{o}_i \right) \textbf{1}^T \right)  & \in \mathbb{R}^{k\times L},   \\
& \boldsymbol{\alpha}_i = \text{softmax} (\textbf{w}^T \textbf{M}_i)   & \in \mathbb{R}^{1\times L},  \\
& \textbf{r}_i = \textbf{Y}_i \boldsymbol{\alpha}^T     & \in \mathbb{R}^k 
\end{align}
where $\textbf{W}^Y$, $\textbf{W}^o \in \mathbb{R}^{k\times k}$ and $\textbf{w}$ are trainable parameters. Vector $\textbf{r}_i$ is the attention-weighted representation of the previous outputs. Thus, the transformed output at time instant with index $i$ is given by:
\begin{align}
& \textbf{o}_i^*=\tanh \left( \textbf{W}^r \textbf{r}_i+ \textbf{W}^x \textbf{o}_i \right)   & \in  \mathbb{R}^k
\end{align} 
where $\textbf{W}^r$, $\textbf{W}^x \in \mathbb{R}^{k\times k}$  are also trainable projection matrices.

The experimental results included in the next section show that attention mechanisms provide a dramatic performance improvement even when dealing with moderately long (length above 4) sequences, which in turn compose the major part of our dataset.

\section{Numerical Experiments}\label{sec:experiments}

The results are obtained using a total of 768 patients for training and 255 for test. Given the nature of the problem to solve, we are more interested in evaluating the recall. K-fold validation is used to validate structural parameters, regularization parameters, and the number of epochs (early stopping). To reduce inter-fold variance, training set in each fold is characterized by roughly a uniform distribution of EHR sequences lengths. I.e., each fold contains the same proportion of sequences of lengths 1, 2, 3, and 4 or more. In Table \ref{tab:params} all the validated parameters for the proposed architecture are included. Following early stopping technique, 100 epochs are chosen for preventing overfitting. Model regularization is achieved by combining $L_2$ regularization of all the weights in the structure and dropout at the input of the final classifier \cite{srivastava2014dropout}.

\begin{table}[]
\centering
\begin{tabular}{@{}ll@{}}
\toprule
\textbf{Parameter}            & \textbf{Value}                   \\ \midrule
RNN hidden units              & {[}clinician: 10, patient: 5{]}     \\
Input NN hidden units         & {[}layer\_1: 10, layer\_2: 20{]} \\
Output NN hidden units        & {[}layer\_1: 10{]}               \\
Initial state NN hidden units & {[}layer\_1: 10, layer\_2: 10{]} \\
Dropout probability           & 0.6                              \\
Epochs                        & 100                              \\
Bath size                     & 20                               \\
Train size                    & 0.75                             \\
Learning rate                 & 0.005                            \\
L2 regularizer factor         & 0.01                             \\ \bottomrule
\end{tabular}
\newline
\caption{Parameter configuration for the model architecture.}
\label{tab:params}
\end{table}

\subsection{Performance discussion}

\begin{table*}[h]
	\centering
	\hspace{0.7cm}
	\begin{tabular}{@{}lllllll@{}}
		\toprule
		\multicolumn{1}{r}{\textbf{}} &\multicolumn{1}{l|}{\textbf{Train}} & \multicolumn{5}{c}{\textbf{Test}} \\ \midrule
		\multicolumn{1}{r}{\textbf{Length}} & All & $1$ & $2$ & $3$ & $\geq 4$ & All \\ \midrule
		NN & 1.14 $\pm$ 0.37 & 0.00 $\pm$ 0.00 & 0.00 $\pm$ 0.00 & 0.00 $\pm$ 0.00 & 0.70 $\pm$ 1.22 & 0.45 $\pm$ 0.43 \\
		LogReg & 65.51 $\pm$ 6.06 & 37.50 $\pm$ 41.46 & 35.41 $\pm$ 20.73 & 43.75 $\pm$ 44.63 & 51.29 $\pm$ 5.12 & 51.24 $\pm$ 5.23 \\
		HMM & 59.92 $\pm$ 6.70 & - & 2.08 $\pm$ 3.62 & 12.50 $\pm$ 21.65 & 51.00 $\pm$ 10.48 & 52.38 $\pm$ 14.62 \\ \midrule
		Linear inputs & 76.43 $\pm$ 5.98 & 25.00 $\pm$ 43.30 & 16.67 $\pm$ 20.41 & 31.25 $\pm$ 44.63 & 58.64 $\pm$ 8.14 & 55.22 $\pm$ 5.66\\
		\textbf{Non-linear inputs} & \textbf{77.30 $\pm$ 3.69} & \textbf{25.00 $\pm$ 43.30} & \textbf{25.00 $\pm$ 21.65} & \textbf{43.75 $\pm$ 44.63} & \textbf{57.78 $\pm$ 4.93} & \textbf{56.14 $\pm$ 4.73} \\ \midrule
		\textbf{Attention, L=1} & \textbf{85.16 $\pm$ 9.92} & \textbf{25.00 $\pm$ 43.30} & \textbf{25.00 $\pm$ 25.00} & \textbf{43.75 $\pm$ 44.63} & \textbf{68.97 $\pm$ 4.08} & \textbf{67.68 $\pm$ 3.50} \\
		Attention, L=2 & 80.70 $\pm$ 3.55 & 25.00 $\pm$ 43.30 & 25.00 $\pm$ 25.00 & 43.75 $\pm$ 44.63 & 64.45 $\pm$ 7.16 & 63.59 $\pm$ 7.10 \\
		Attention, L=3 & 81.63 $\pm$ 5.02 & 25.00 $\pm$ 43.30 & 25.00 $\pm$ 25.00 & 43.75 $\pm$ 44.63 & 66.69 $\pm$ 3.60 & 65.57 $\pm$ 3.12 \\
		Attention, L=4 & 80.76 $\pm$ 7.47 & 25.00 $\pm$ 43.30 & 25.00 $\pm$ 25.00 & 43.75 $\pm$ 44.63 & 72.63 $\pm$ 7.53 & 71.16 $\pm$ 7.04 \\ \bottomrule
	\end{tabular}%
	\newline
	\caption{ Recall provided by different models, including variations of our proposal. We test the model for every group of EHR sequences (depending on the length). All the values correspond to the mean and standard deviation of the metric obtained with 4 folds. Decision threshold is selected for each case by cross-validation.  
	}
	\label{tab:results}
\end{table*}

Table \ref{tab:results} shows recall average performance and standard deviation measured along the different folds of three baseline models and five different variations of our model. Test performance is stratified, considering clinician EHR sequences of length 1, 2, 3, and 4 or more. However, we note that the vast majority of suicidal ideation cases (more specifically, 65.43\% of true positives) in the dataset are contained in long EHR sequences (4 or more). Namely, achieving a good performance in the last group is crucial.

With NN we refer to a feed-forward neural network of two layers with 10 and 20 hidden units, respectively. LogReg corresponds to a simple logistic regression with $L_2$ norm regularization, which is one of the simplest classifiers. These two cases are trained using the stacked vectors of visits $\textbf{x}_{n, i}^c$ and the most recent answers $\textbf{x}_{n, j}^p$, i.e., sequential nature is ignored. HMM is a first order Hidden Markov Model with both Gaussian and labels emission and 5 hidden states.

With the fourth and fifth rows we can compare the effect in our method of using a linear transformation, versus a pre-trained dense NN layer to perform dimensionality reduction at the RNN input (see Section \ref{sec:seq_mod}). It can be observed that the use of a pre-trained dense NN layer, which already provides discriminative features to the RNN, provides both better train and test performance, particularly for short sequences. The last four rows correspond to the model versions that use attention-based patient RNN outputs, with attention windows varying from $L=1$ to $L=4$. The version with $L=1$ is chosen as the most suitable, considering a trade-off between recall performance, computational cost, and results' standard deviation. Note that for $L=1$ the patient RNN contains features related to the variation between two consecutive RNN states. Observe the remarkable performance improvement in this latter case, particularly for long clinician sequences, which correspond to patients that have been under psychiatric treatment for a sufficiently long time. 

For short EHR sequences, all models provide poor recall and large inter-fold standard deviation. This is certainly related to the lack of data of this type in our dataset. Note however that the use of non-linear RNN inputs in our model helps to cope with EHR sequences with length equal to two. Even without attention, our proposal clearly improves the baselines (NN, LogReg, and HMM) for long (length 3 or more) EHR sequences. Both Logistic Regression and HMM are comparable (they both achieve roughly a recall of 51\% for long sequences), while NN certainly is not able to predict the suicidal ideation cases. Our model with attention and $L=1$ at the patient sequence achieves a recall close to 67\% with the smallest variance across folds. For this model, other metrics evaluations are included in table \ref{tab:metrics} for the chosen version.

\begin{table*}[]
	\centering
	\hspace{0.9cm}
	\begin{tabular}{@{}lllllll@{}}
		\toprule
		\multicolumn{1}{r}{\textbf{}} & \multicolumn{1}{l|}{\textbf{Train}} & \multicolumn{5}{c}{\textbf{Test}} \\ \midrule
		\multicolumn{1}{r}{\textbf{Length}}  & All & $1$ & $2$ & $3$ & $\geq 4$ & All \\ \midrule
		Accuracy & 95.98 $\pm$ 0.60 & 92.65 $\pm$ 9.15 & 91.26 $\pm$ 5.91 & 91.00 $\pm$ 9.38 & 88.76 $\pm$ 5.11 & 88.96 $\pm$ 5.04 \\
		Precision & 87.08 $\pm$ 6.63 & 25.00 $\pm$ 43.30 & 33.18 $\pm$ 34.82 & 19.33 $\pm$ 21.05 & 58.88 $\pm$ 5.11 & 58.81 $\pm$ 20.06  \\
		Recall & 85.16 $\pm$ 9.92 & 25.00 $\pm$ 43.30 & 25.00 $\pm$ 25.00 & 43.75 $\pm$ 44.63 & 68.97 $\pm$ 4.08 & 67.68 $\pm$ 3.50 \\
		AUC & 94.02 $\pm$ 4.22 & 58.65 $\pm$ 38.40 & 67.33 $\pm$ 34.82 & 59.07 $\pm$ 37.48 & 83.85 $\pm$ 21.05 & 83.29 $\pm$ 5.13 \\
		F1-score & 85.34 $\pm$ 4.02 & 25.00 $\pm$ 43.30 & 28.43 $\pm$ 28.62 & 26.11 $\pm$ 26.68 & 61.83 $\pm$ 11.81 & 61.33 $\pm$ 4.03\\ \bottomrule
	\end{tabular}%
\newline
\caption{Performance metrics of the model proposed with $L=1$ attention.}
\label{tab:metrics}
\end{table*}

\begin{figure}
	\centering
	\includegraphics[width=0.9\linewidth]{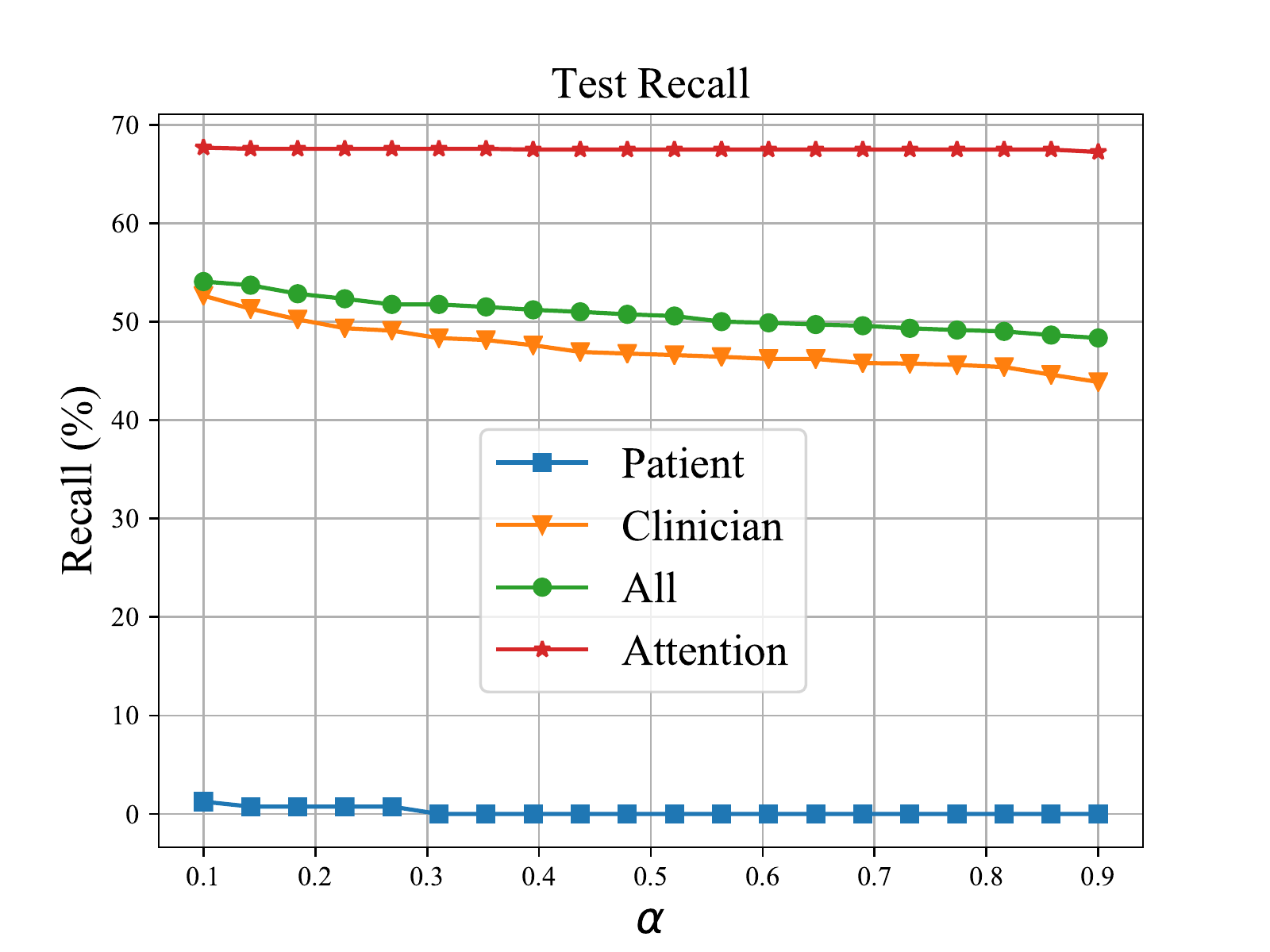}
	\caption{Test recall results evaluating the merging effect of the sequences, depending on the threshold for detection. The lines represent the results obtained using only the patient (blue) or the clinician (orange) sources, using both (green) and using both with attention schemes (red).}
	\label{fig:patient_vs_doctor}
\end{figure}

The results in Table \ref{tab:results} demonstrate that EMA is a significant support for automatically detect suicidal ideation when is used along with the clinician information. Results indeed demonstrate that inspecting the contrast between two consecutive questionnaires rather than basing the prediction exclusively on the latest sets of responses improves the performance. This information is exactly what the attention model is attempting to capture. A similar conclusion is drawn when we compare the performance of the model when either clinician RNN, or the patient RNN are not used (all parameters are set to 0 and the RNN is not trained so it does not influence the classifier). Fig. \ref{fig:patient_vs_doctor} shows the recall achieved by these models as a function of the classifier threshold $\alpha$. Not surprisingly, when the clinician RNN is removed and only the EMA information is used, the model is no longer valid to predict suicidal ideation diagnose, as the EMA sequence is not aligned in time and completely independent of the suicidal ideation identification sequence. However, when both sources of information are used and attention is used to process the patient RNN prediction recalls improves in close to a 20\% w.r.t. to the model when the patient RNN is not used or only the latest Memind questionnaire is considered.

\subsection{Most relevant features}

After model has been validated in terms of test performance, we can study what input features in both the clinician sequence and patient sequence impact the most in the classifier’s prediction. To this end, we analyze the profile of the weights in the first layer of the NN that performs the input dimensionality reduction. Thus, we have a set of 93 values associated to each of the clinician input features and 24 for the patient features. Tables \ref{tab:top_ten_doctor} and \ref{tab:top_ten_patient} shows the top ten for the clinician and patient sources, respectively. We also indicate the associated normalized weight (absolute
value of the weight divided by the maximum). Observed that, while in the clinician sequence we have four input features that are clearly dominant, a similar effect is not present in the patient sequence. This is indeed expected, since we have observed that the patient sequence really matters when two consecutive questionnaires are jointly processed using the attention mechanism. It is important to highlight that two questions out of the total of five directly related to suicidal risk were not found within the dominant input features of the clinician sequence, these are suicidal plans and suicide attempt since the last visit. This finding is consistent with previous studies that criticize the traditional temporal sequence of stages in the suicidal process \cite{baca2011estimating}. Furthermore, it is interesting to remark how diagnoses less commonly associated with suicidal ideation such as F00-F09 diagnoses are found to be of greater importance compared to those most frequently studied (F40-F49 and F30-F39 ICD-10 codes). Although not shown in Table III, it is worth to mention that the 12-th question among the largest weights corresponds to the unemployed item in Brugha's life threatening events scale, which is a common clinical finding. In EMA sequence, questions related to despair, aggression, confidence or rest have largest weights than more specific questions like “Have you ever felt that you had no desire to live?”, which is in number 13 of the 24 input features. This is of clinical interest, as indirect and non-intrusive questions can make clinicians be alert about suicidal thoughts in patients.

\begin{table}[]
\centering
\begin{tabular}{@{}ccl@{}}
\toprule
\multicolumn{3}{c}{\textbf{Clinician}} \\ \midrule
\multicolumn{1}{c}{\textbf{Index}} & \multicolumn{1}{c}{\textbf{Score}} & \multicolumn{1}{c}{\textbf{Feature}} \\ \midrule
1 & 1.00 & Have you ever thought about harming yourself? \\
2 & 0.81 & \begin{tabular}[c]{@{}l@{}}Have you ever thought you would be better \\off dead or wish you were dead?\end{tabular} \\
3 & 0.65 & Have you ever attempted suicide? \\
4 & 0.51 & Unemployed with benefit. \\
5 & 0.48 & Living with siblings. \\
6 & 0.45 & Shared residence. \\
7 & 0.44 & \begin{tabular}[c]{@{}l@{}}F00-F09 Mental disorders due to known \\physiological conditions.\end{tabular} \\
8 & 0.41 & Electroconvulsive therapy. \\
9 & 0.40 & Living with offspring. \\
10 & 0.40 & Temporary disability. \\ \bottomrule
\end{tabular}%
\newline
\caption{Top ten more relevant features from the clinician (EHR) source.}
\label{tab:top_ten_doctor}
\end{table}

\begin{table}[]
\centering
\begin{tabular}{@{}ccl@{}}
\toprule
\multicolumn{3}{c}{\textbf{Patient}} \\ \midrule
\textbf{Index} & \textbf{Score} & \multicolumn{1}{c}{\textbf{Feature}} \\ \midrule
1 & 1.000 & \begin{tabular}[c]{@{}l@{}}Have you recently felt you could no overcome your\\ difficulties? (GHQ-12, item 6)\end{tabular} \\
2 & 0.98 & Anger, arguments or fights. \\
3 & 0.92 & \begin{tabular}[c]{@{}l@{}}Have you recently been thinking of yourself as a \\ worthless person? (GHQ-12, item 11)\end{tabular} \\
4 & 0.91 & \begin{tabular}[c]{@{}l@{}}I have felt active and vigorous (WHO-5, \\ item 3)\end{tabular} \\
5 & 0.91 & \begin{tabular}[c]{@{}l@{}}Have you recently felt capable of making decisions\\ about things (GHQ-12, item 4)\end{tabular} \\
6 & 0.90 & \begin{tabular}[c]{@{}l@{}}Have you recently been losing confidence in\\ yourself? (GHQ-12, item 10)\end{tabular} \\
7 & 0.90 & \begin{tabular}[c]{@{}l@{}}Have you recently been feeling reasonably happy,\\ all things considered? (GHQ-12, item 12)\end{tabular} \\
8 & 0.89 & \begin{tabular}[c]{@{}l@{}}I woke up feeling fresh and rested (WHO-5,\\ item 4)\end{tabular} \\
9 & 0.89 & \begin{tabular}[c]{@{}l@{}}Have you recently felt constantly under strain?\\ (GHQ-12, item 5)\end{tabular} \\
10 & 0.88 & \begin{tabular}[c]{@{}l@{}}Have you recently been able to enjoy your normal\\ day to day activities? (GHQ-12, item 7)\end{tabular} \\ \bottomrule
\end{tabular}
\newline
\caption{Top ten more relevant features from the patient (EMA) source.}
\label{tab:top_ten_patient}
\end{table}

\subsection{Latent Space Visualization}

\begin{figure*}[]
	\centering
	\setcounter{subfigure}{0}
	\subfigure[Real suicidal ideation]{\includegraphics[width=0.44\linewidth]{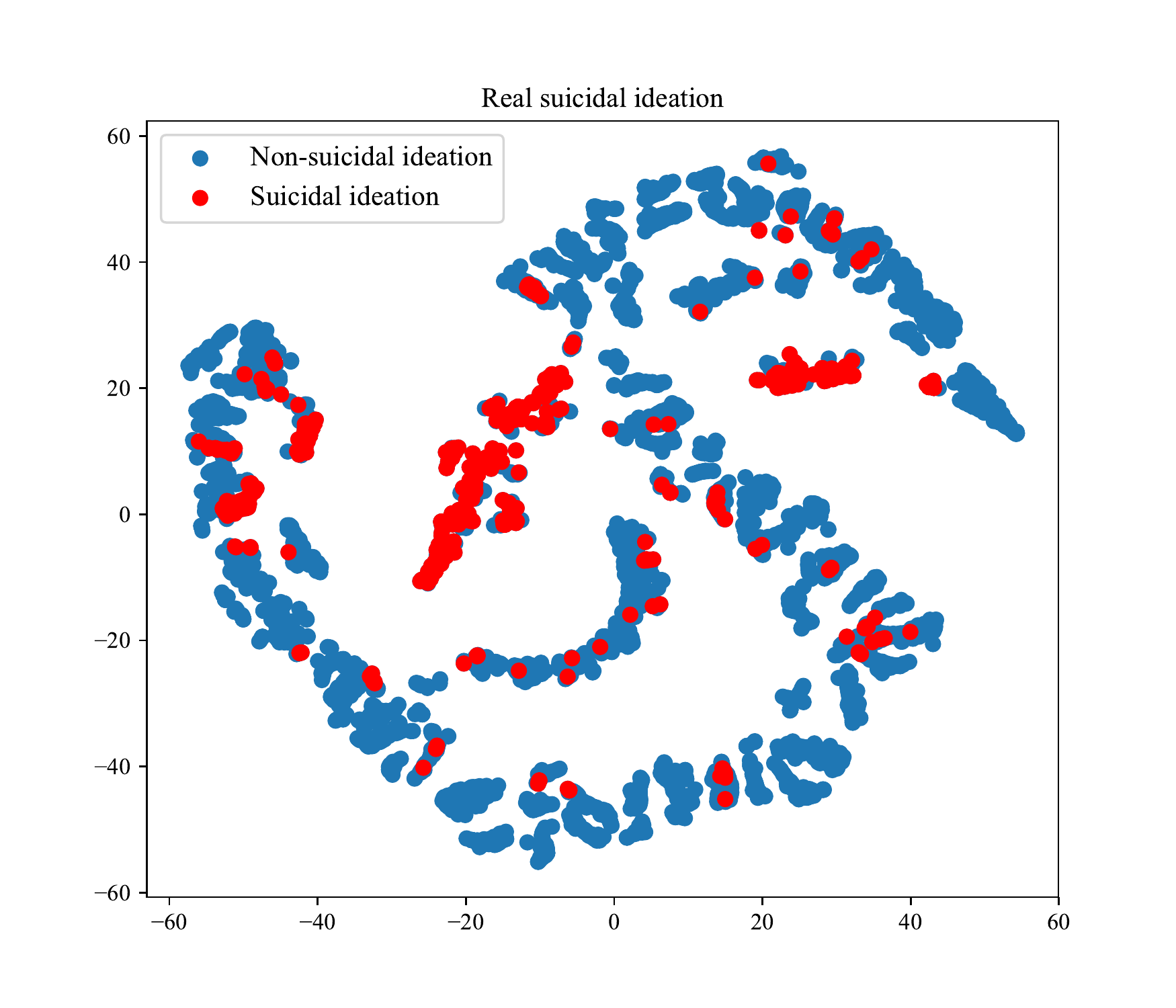}}
	\subfigure[Suicidal ideation probability]{\includegraphics[width=0.539\linewidth]{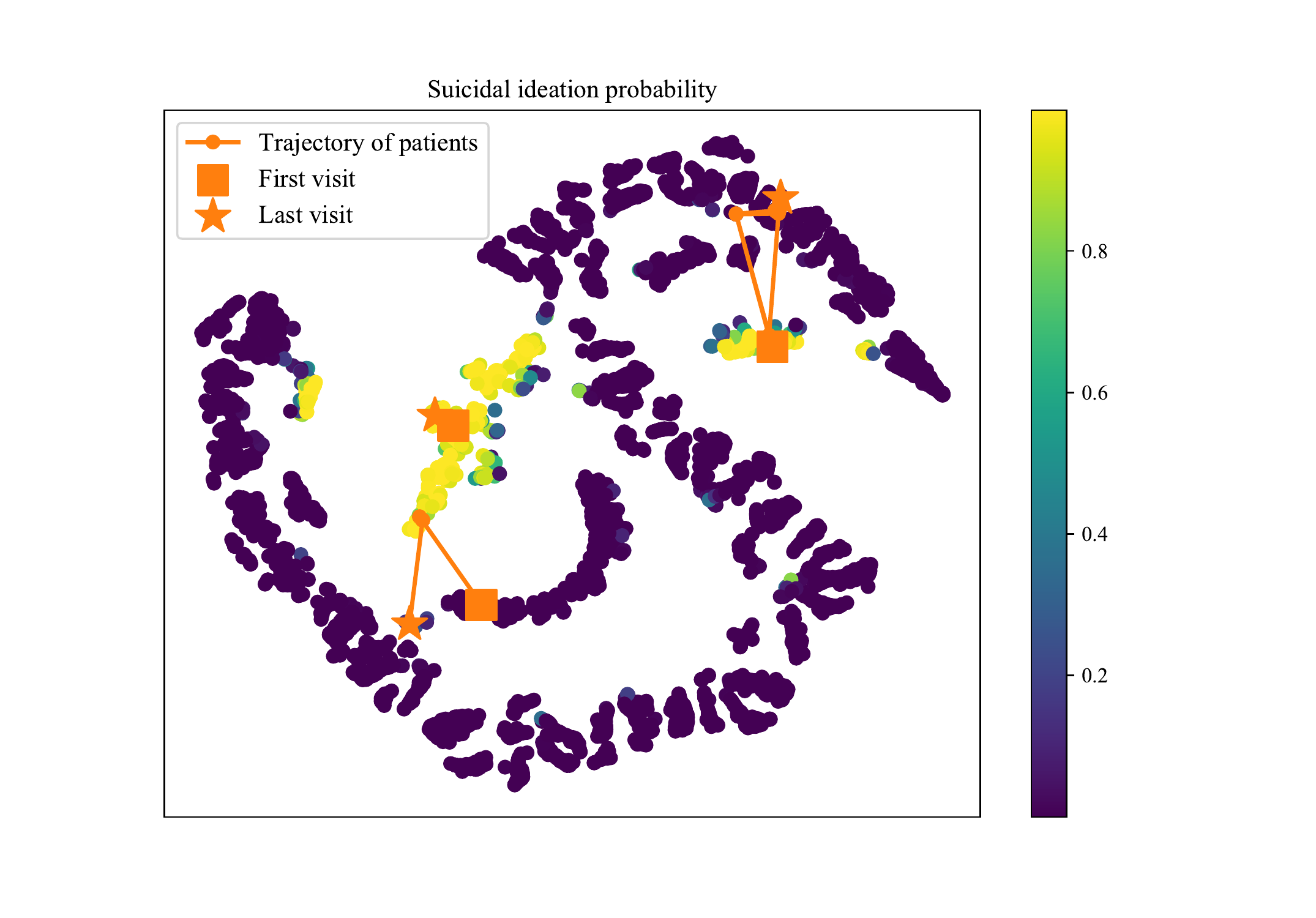}}
	\setcounter{subfigure}{0}
	\caption{Dimensionality reduction of the latent space (the vector that concatenates $\textbf{o}_i^c$, $\textbf{o}_{j}^p$ and $\textbf{x}^s$ in Figures \ref{fig:diag_normal} and \ref{fig:diag_att}), using t-SNE. Each point corresponds to a visit. Left: original suicidal ideation. Right: the color map is the probability of suicidal ideation (obtained logits of the final feed forward network). Indeed, trajectories in the map for three different patients are included.  }
	\label{fig:tsne}
\end{figure*}

One of the main advantages of deep learning methods is the ability of implicitly finding latent representation of the data. For that purpose we are employe t-SNE \cite{maaten2008visualizing} (t-Distributed Stochastic Neighbor Embedding), a powerful technique for visualizing high-dimensional data using a transformation that reveals structure at many different scales. We are applying this technique on the vectors that feed the final classifier to evaluate the capacity of discrimination of the RNN networks. For all the experiments the t-SNE perplexity parameter is fixed to 100 and the learning rate to 10. In Figure \ref{fig:tsne} we include two examples of t-SNE visualizations of the outputs of the RNNs. At the left we include the observed suicidal ideation corresponding to each point (i.e. the label we aim at predicting), while at the right we plot the probability of suicidal ideation that the model obtains. Indeed, we include trajectories in the map from three different patients. It seems clear than the t-SNE transformation reveals clouds of points with groups of patients that present suicidal ideation, and the state `travels’ between them depending on the evolution of the suicidal ideation of the patient. As a by product, observe that in the right map all positives are concentrated and do not overfit to the isolated positive targets that can be seen in the left map. This is a sign that the model is not overfitting the training data.

\section{Discussion}\label{sec:discussion}

A model for asynchronous, length-variable, randomly sampled sequences with heterogeneous data has been developed, using RNNs for the sequential inputs with a first non linear transformation, given by pre-trained feed-forward NNs, to cope with the heterogeneity of the data and perform a dimensionality reduction. The analysis of the input layer of the feed-forward networks has allowed us to obtain a measure of the features relevance, and perform clinical interpretations. Furthermore, attention mechanisms are incorporated to the \emph{patient sequence} for addressing the variability in sequence lengths. We have applied a merging technique consisting on temporally aligning the outputs of each RNN according to time labels. This merged vectors feed another feed forward NN that predicts the suicidal ideation at each visit to the clinician. In our case, the suicidal ideation sequence imposed the date and we extracted from each RNN the output at last time instant before that date. We have built a cost function that takes the sum of the cost in all visits of a patient. We have demonstrated that this merging procedure improves the performance of the model, rather than if each sequence was used separately.

Our model predicts suicidal ideation interpreted by a clinician, based on the past mental health visits and the history of answers of a patient collected through a web-based EMA system, that provides a promising complementary tool for the clinician. The visualization step has allowed us to discover the trajectories of the patient within the latent space, observing that there exist some clouds with positive and negatives samples, and there exists a coherence between the sequential target of the patient and the trajectory on the t-SNE map.

This is, to our knowledge, a novel study with two main innovations: 1 the combination of Electronic Health Records and Ecological Momentary Assessment data for predicting suicidal ideation, and 2 the use of Deep Learning sequential architectures (RNNs) to discover temporal correlations in psychiatric patients' evolution. We have demonstrated that our method outperforms other existing methods in the literature. Further, we have exposed an interesting feature of our model: the interpretability of the latent space via dimensionality reduction techniques as t-SNE. 

Future work could be centered on including more sequences associated to the patient that can be merged and aligned, for instance, the complete Electronic Health Record of the patient (including primary care visits). Another branch of the research will be the application of more complex attention schemes on every sequence, with different temporal windows. At last, we will research in such an interesting application of this work consisting on generative models for this medical data.

\hfill

\ifCLASSOPTIONcaptionsoff
  \newpage
\fi

\bibliographystyle{IEEEtran}
\bibliography{IEEEabrv,biblio}

\newpage

\end{document}